%% file: paper.tex
\documentclass{article}
\usepackage{spconf,amsmath,epsfig, amssymb}

\usepackage[utf8]{inputenc}
\usepackage{pgfplots}

\usepgfplotslibrary{external}
\usepgfplotslibrary{groupplots}

\usetikzlibrary{shapes.geometric}
\usepackage{contour}

\usepackage{tikz}

\usepackage{caption}
\usepackage{subcaption}

\usepackage{blindtext}

\usepackage{booktabs}
\usepackage{siunitx}
\DeclareSIUnit\pixel{px}
\usepackage{tabularx}
\usepackage{multirow}
\usepackage{xcolor,colortbl}
\usepackage{etoolbox}
\robustify\cellcolor
\usepackage{array} 

\definecolor{clc112}{RGB}{255, 000, 000}
\definecolor{clc121}{RGB}{204, 077, 242}
\definecolor{clc122}{RGB}{204, 000, 000}
\definecolor{clc133}{RGB}{255, 077, 255}
\definecolor{clc211}{RGB}{255, 255, 168}
\definecolor{clc231}{RGB}{230, 230, 077}
\definecolor{clc311}{RGB}{128, 255, 000}
\definecolor{clc312}{RGB}{000, 166, 000}
\definecolor{clc313}{RGB}{077, 255, 000}
\definecolor{clc512}{RGB}{128, 242, 230}

\DeclareUnicodeCharacter{2212}{−}
\usepgfplotslibrary{groupplots,dateplot}
\usetikzlibrary{patterns,shapes.arrows}
\pgfplotsset{compat=newest}

\title{SeasoNet: A Seasonal Scene Classification, segmentation and Retrieval dataset for satellite Imagery over Germany}
\name{Dominik Ko\ss{}mann\textsuperscript{1,$\dagger$}, Viktor Brack\textsuperscript{1,$\dagger$}, Thorsten Wilhelm\textsuperscript{2}}
\address{Pattern Recognition in Embedded Systems Group\textsuperscript{1}, Image Analysis Group\textsuperscript{2},
	\\ TU Dortmund University, Germany}

\begin{document}

\maketitle

\begin{abstract}

This work presents SeasoNet, a new large-scale multi-label land cover and land use scene understanding dataset. It includes $1\,759\,830$ images from Sentinel-2 tiles, with 12 spectral bands and patch sizes of up to $ 120 \ \mathrm{px} \times 120 \ \mathrm{px}$. Each image is annotated with large scale pixel level labels from the German land cover model LBM-DE2018 with land cover classes based on the CORINE Land Cover database (CLC) 2018 and a five times smaller minimum mapping unit (MMU) than the original CLC maps. We provide pixel synchronous examples from all four seasons, plus an additional snowy set. These properties make SeasoNet the currently most versatile and biggest remote sensing scene understanding dataset with possible applications ranging from scene classification over land cover mapping to content-based cross season image retrieval and self-supervised feature learning. We provide baseline results by evaluating state-of-the-art deep networks on the new dataset in scene classification and semantic segmentation scenarios. 

\end{abstract}
\vspace{-0.8em}
\section{Introduction}
\label{sec:intro}
Automatic{\let\thefootnote\relax\footnotetext{$\dagger$ Authors contributed equally}} Earth monitoring and remote sensing applications produce a huge amount of unlabeled data every day. Fast analysis of this data is essential to land use and climate change monitoring as well as disaster prevention. An analysis includes solving various vision tasks to understand a satellite image scene to the fullest. Tasks range from land cover classification over image retrieval to semantically mapping each pixel. Further, all areas need to be agnostic to seasonal changes. Therefore, most approaches leverage deep learning architectures which need a huge amount of labeled data to train directly on the target remote sensing (RS) domain. Transfer learning scenarios between the RS domain and natural scene images with pre-trained features from ImageNet unfortunately fall short in performance~\cite{Manas2021SeasonalContrast}. While some large-scale RS benchmark datasets for scene classification (e.g., BigEarthNet~\cite{Bigearth_1} or SEN12MS~\cite{sen12ms_dataset_2019}) exist, they do not offer pixel level labels or only provide small scale segmentation, see Table~\ref{tab:better_resolution}. This limits the potential of change detection and semantic land cover mapping. Furthermore, seasonal changes can introduce a significant domain shift. Current datasets include instances from multiple seasons only in different locations and without pixel labels. Thus, the domain shift can not be considered in image retrieval or segmentation scenarios.

\newcommand{\samplesize}{0.0925}
\newcommand{\parboxsize}{1.0cm}
\begin{table}[t!]
	\resizebox{1\linewidth}{!}{%
		\renewcommand{\arraystretch}{0.6}   
		\begin{tabular}{m{2.2cm}m{2.8cm}m{2.2cm}m{2.2cm}}
			\toprule
			\multicolumn{1}{c}{\textbf{Sentinel-2}} & \multicolumn{1}{c}{\textbf{Multi-Label~\cite{Bigearth_1}}} & \multicolumn{1}{c}{\textbf{Pixel-Label~\cite{wilhelm2021land}}} & \multicolumn{1}{c}{\textbf{This Work}}\\
			\midrule
			
			\includegraphics[width=\samplesize\textwidth]{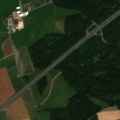} \centering
			&
			\small{
			\textcolor{clc311}{Broad-leaved forest},
			\textcolor{clc211}{Non-irrigated arable land},
			\textcolor{clc231}{Pastures}
			}
			&
			\includegraphics[width=\samplesize\textwidth]{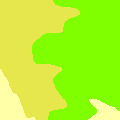} \centering
			&
			\includegraphics[width=\samplesize\textwidth]{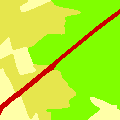} \centering
			\tabularnewline
			
			\includegraphics[width=\samplesize\textwidth]{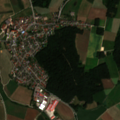} \centering
			&
			\small{
			\textcolor{clc312}{Coniferous forest},
			\textcolor{clc112}{Discontinuous urban fabric},
			\textcolor{clc211}{Non-irrigated arable land},
			\textcolor{clc231}{Pastures}
			}
			&
			\includegraphics[width=\samplesize\textwidth]{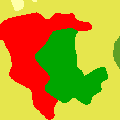} \centering
			&
			\includegraphics[width=\samplesize\textwidth]{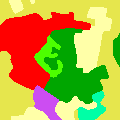} \centering
			\tabularnewline
			
			\bottomrule 
	\end{tabular}}
	\caption{Overview of current land cover dataset resolutions and labels in comparison to our dataset. In contrast to image level labels~\cite{Bigearth_1} or pixel-level labels~\cite{wilhelm2021land} this work adopts the land-cover and land-use maps of the German land cover model LBM-DE2018, which has a higher mapping resolution and therefore misses significantly less small objects.}
	\label{tab:better_resolution}
	\centering
	\vspace{-1.5em}
\end{table}

\begin{table*}[t!]
	\renewcommand{\arraystretch}{0.6}   
	\begin{tabular}{m{5.8cm}m{1.5cm}m{1.5cm}m{1.5cm}m{1.5cm}m{1.5cm}m{1.5cm}}
		\toprule
		\multicolumn{1}{c}{\textbf{Image-level labels}} & \multicolumn{1}{c}{\textbf{Pixel-level}} & \multicolumn{1}{c}{\textbf{Spring}} & \multicolumn{1}{c}{\textbf{Summer}} & \multicolumn{1}{c}{\textbf{Fall}} & \multicolumn{1}{c}{\textbf{Winter}} & \multicolumn{1}{c}{\textbf{Snow}}\\
		& \multicolumn{1}{c}{\textbf{labels}}		& \multicolumn{1}{c}{($ 460\,104 $)} 	  & \multicolumn{1}{c}{($ 481\,456 $)} 		& \multicolumn{1}{c}{($ 500\,149 $)} 	& \multicolumn{1}{c}{($ 218\,663 $)} 	  & \multicolumn{1}{c}{($ 99\,458 $)}\\
		\midrule
		
		\textcolor{clc122}{Road and rail networks and associated land},
		\textcolor{clc211}{Non-irrigated arable land},
		\textcolor{clc231}{Pastures},
		\textcolor{clc311}{Broad-leaved forest}
		&
		\includegraphics[width=\samplesize\textwidth]{img/examples/49199220_10128145_label.png} \centering
		&
		\includegraphics[width=\samplesize\textwidth]{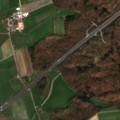} \centering
		&
		\includegraphics[width=\samplesize\textwidth]{img/examples/49199220_10128145_10m_RGB_summer.png} \centering
		&
		\includegraphics[width=\samplesize\textwidth]{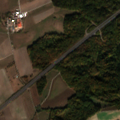} \centering
		&
		\includegraphics[width=\samplesize\textwidth]{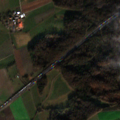} \centering
		&
		\includegraphics[width=\samplesize\textwidth]{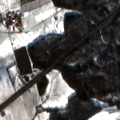} \centering
		\tabularnewline
		
		\textcolor{clc112}{Discontinuous urban fabric},
		\textcolor{clc121}{Industrial or commercial units},
		\textcolor{clc211}{Non-irrigated arable land},
		\textcolor{clc231}{Pastures},
		\textcolor{clc312}{Coniferous forest},
		\textcolor{clc313}{Mixed forest},
		\textcolor{clc512}{Water bodies}
		&
		\includegraphics[width=\samplesize\textwidth]{img/examples/49262496_11314703_label.png} \centering
		&
		\includegraphics[width=\samplesize\textwidth]{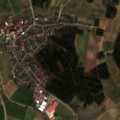} \centering
		&
		\includegraphics[width=\samplesize\textwidth]{img/examples/49262496_11314703_10m_RGB_summer.png} \centering
		&
		\includegraphics[width=\samplesize\textwidth]{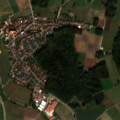} \centering
		&
		\includegraphics[width=\samplesize\textwidth]{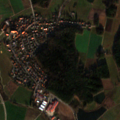} \centering
		&
		\includegraphics[width=\samplesize\textwidth]{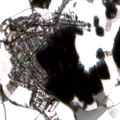} \centering
		\tabularnewline
		
		\textcolor{clc112}{Discontinuous urban fabric},
		\textcolor{clc122}{Road and rail networks and associated land},
		\textcolor{clc133}{Construction sites},
		\textcolor{clc231}{Pastures},
		\textcolor{clc311}{Broad-leaved forest},
		\textcolor{clc312}{Coniferous forest},
		\textcolor{clc313}{Mixed forest}
		&
		\includegraphics[width=\samplesize\textwidth]{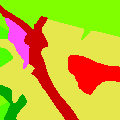} \centering
		&
		\includegraphics[width=\samplesize\textwidth]{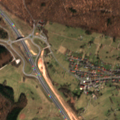} \centering
		&
		\includegraphics[width=\samplesize\textwidth]{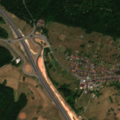} \centering
		&
		\includegraphics[width=\samplesize\textwidth]{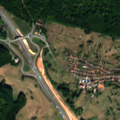} \centering
		&
		\includegraphics[width=\samplesize\textwidth]{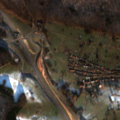} \centering
		&
		\includegraphics[width=\samplesize\textwidth]{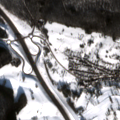} \centering
		\tabularnewline
		
		\bottomrule 
	\end{tabular}
	\caption{Example of the Sentinel-2 RGB channels in SeasoNet for four given seasons and the snow set. 
	}
	\label{tab:season_examples}
	\centering
	\vspace{-1.5em}
\end{table*}

In this work, we introduce SeasoNet\footnote{https://doi.org/10.5281/zenodo.5850307}, a new large-scale multi-spectral multi- and pixel-label remote sensing benchmark dataset. 
It consists of $1\,759\,830$ Sentinel-2 image patches, annotated with multi-label land cover and land usage classes from the CORINE Land Cover database (CLC) 2018\footnote{https://land.copernicus.eu/pan-european/
corine-land-cover/clc2018}, covering the total area of Germany. To the best of our knowledge it is significantly larger than the currently biggest labeled multi-spectral archives (see Table~\ref{tab:dataset_comparison}). Further, we  provide large scale pixel level labels based on these land cover classes. These annotations are adopted from a 5 hectare MMU version of the CLC database from the publicly available German land cover model LBM-DE2018\footnote{https://gdz.bkg.bund.de/index.php/default/catalog/product/view/id/1071 \\ /s/corine-land-cover-5-ha-stand-2018-clc5-2018/}. It offers a 5 times better minimum object resolution than the original CLC database with 25 hectares. We offer individual sets for the four seasons and an additional snowy set.
We believe this dataset serves well as a resource for the development and evaluation of new models in scene classification, mapping and retrieval. It may also serve as a basis for transfer-learning to other data in the RS domain or pre-training of features in a self-supervised setup. 
The dataset will be made publicly available alongside code for data preparation and evaluation.

\vspace{-1em}
\section{Challenges in Remote Sensing Scene Understanding}
\label{sec:challenges}
\vspace{-0.8em}
One of the main challenges in remote sensing scene understanding is the huge amount of data necessary to learn domain agnostic features with modern deep architectures. For natural scene images supervised pre-training with ImageNet~\cite{deng2009imagenet} is well established. Unfortunately, in a transfer learning setup the generated features do not always generalize well to new domains like remote sensing imagery. Furthermore, RS imagery often contains multi or hyper-spectral images, which do not match features learned from ImageNet RGB channels, thus making the domain gap more severe.
There are currently multiple large-scale archives to learn in domain features (Table~\ref{tab:dataset_comparison}), but these still do not match the size of ImageNet~\cite{deng2009imagenet} with its 1,2 million images.

Another issue are seasonal changes to vegetation and weather, which create a domain shift in the appearance. In current remote sensing archives, e.g., SEN12MS~\cite{sen12ms_dataset_2019} or SeasonalContrast~\cite{Manas2021SeasonalContrast}, multiple seasons are already included, but they are sampled over different locations and thus learning seasonal changes on an object level becomes difficult.
Current works try to tackle seasonal changes by e.g., style transfer with generative adversarial networks~\cite{li2021image}. But, it remains a challenge to content-based image retrieval~\cite{li2021image} as well as change detection and classification tasks~\cite{Manas2021SeasonalContrast,OSCDdaudt2018}. 

For land cover mapping and change detection on pixel level not many large-scale datasets with a diverse set of classes and a non-local coverage exist so far, see Table~\ref{tab:dataset_comparison}.
In addition, they can suffer from erroneous labels, since creating a grid over a given annotated land cover map to create image patches can result in small areas at the edges of images~\cite{wilhelm2021land}. These areas are impossible to segment on a pixel level or classify on an image level without the necessary context information. Thus, this problem affects scene classification and segmentation datasets alike. Moreover, the underlying mapping resolution might not fit the given satellite image resolution, making false pixel labels at the borders of regions likely.
\vspace{-0.8em}
\section{The SeasoNet Dataset}
\label{sec:dataset}
\vspace{-0.8em}
Our dataset aims to tackle the above RS challenges. SeasoNet consists of $1\,759\,830$ multi-spectral multi- and pixel-label image patches from the Sentinel-2 mission, covering the whole area of Germany. 
The dataset is constructed from 311 Sentinel-2 tiles covering Germany, acquired between April 2018 and February 2019. We use the same 12 spectral bands from Sentinel-2 as \cite{Bigearth_1} with Level-2A Bottom-of-Atmosphere correction. Two sets of patches were created from two regular grids over the selected tiles. A singular grid consists of non overlapping connected patches. The two grids are shifted by half the patch size in both dimensions and thus overlap.
By this process we were able to sample different large scale regions, since each grid avoids different small scale cut off regions at image borders. 
Each patch includes sizes of $ 120 \ \mathrm{px} \times 120 \ \mathrm{px}$, $ 60 \ \mathrm{px} \times 60 \ \mathrm{px}$, $ 20 \ \mathrm{px} \times 20 \ \mathrm{px}$ for $ 10 \ \si{\meter}$, $ 20 \ \si{\meter}$, $ 60 \ \si{\meter}$ Sentinel-2 bands, respectively. In total, we sample from $ 519\,547 $ unique patch locations. 
The acquisition of images per patch location has been split by season into four sets plus an extra snowy set. Season date boundaries are based on their meteorological definitions, with an added gap of one month between them, ensuring that each image is representative
for its season.
All seasons except winter include only images with less than $1\%$ snow and less than $5\%$ clouds. For winter these thresholds were both set to $10\%$, because
of the high confusion rate between frost, snow and clouds. The minimum snow amount of the snowy set is also $10\%$, aligning with the maximum threshold during winter. A fixed
maximum value for clouds was not set on the snowy images, instead a classifier
was trained to specifically find and remove cloudy images, with a focus on high recall of cloudy images.
After preprocessing, there are $ 181\,480 $ sample locations for which an image could be found in each season, thus a subset of $ 181\,480 \ \mathrm{locations} \cdot4\ \mathrm{seasons} = 725\,920 $ dataset samples has full season coverage. 
For a total of $88\%$ of all patch locations we found images for at least three seasons.
The snowy set consists of $ 99\,458 $ images, see Table~\ref{tab:season_examples}. Thus, it is possible to track seasonal changes over the same patch location while providing visually varying and diverse scenes. 

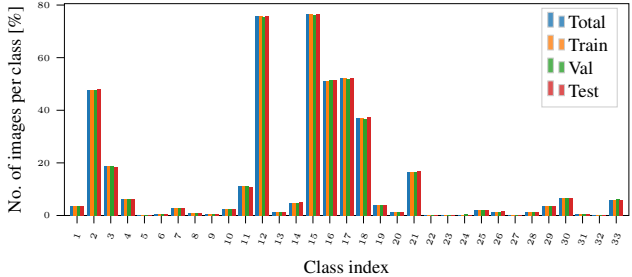
\begin{figure}[t!]
	\centering
	\tikzsetnextfilename{split_classes_grid2}%
	\input{split_classes_grid2.tex}%

	\caption{Histogram of relative class distribution for the total number of $ 879\,374 $ samples and the 70\% training/20\% test/10\% validation split partitioning for grid 2 of SeasoNet. The class distribution over the individual splits is nearly identical and thus provides a good basis for evaluation.}
	\label{fig:split_classes}
	\vspace{-1em}
\end{figure}

In Germany 35 of the 43 level-3 CLC classes are present.
Two classes (\textit{Glaciers and perpetual snow, Burnt Areas}) are additionally omitted due to an insufficient number of samples. As a result, 33 classes are included in SeasoNet. 
The number of samples per land cover class varies significantly over the dataset, thus making it a challenging but realistic RS scene understanding dataset (cf. Fig.~\ref{fig:split_classes}). Additionally, we included the level of urbanization inside the metadata of each patch, provided by the GE250 region classification map\footnote{http://gdz.bkg.bund.de/index.php/default/gebietseinheiten-1-250-000-ge250.html}, enabling further dataset splits for e.g., retrieval tasks.

Our large-scale dataset helps in tackling the transfer learning domain gap and improving land cover classification tasks. It further provides a benchmark for understanding seasonal changes not only on an image but also on a mapping level of individual objects, thus supporting the evaluation and creation of new methods for style transfer and content-based retrieval methods. The high resolution land cover pixel annotations avoid problems with small cut off regions and the sampling over two grids enables more context information being represented inside the dataset, avoiding the difficulties described in~\cite{wilhelm2021land}. Because of this, learning a segmentation of small regions like streets or railways on a huge dataset becomes possible. Even self-supervised feature learning methods relying on seasonal changes~\cite{Manas2021SeasonalContrast} now are possible on an object~\cite{Van_Gansbeke_2021_ICCV} and not only an image level. 

\begin{table}[t!]
	\resizebox{1\linewidth}{!}{%
		\begin{tabular}{lccr}
			\toprule
			Dataset name & Image Type & Annotation Type & Number of Images \\
			\midrule
			Agriculture-Vision~\cite{chiu2020agriculture} & Aerial Multispectral & Multi-Label + Pixel Label & $ 94\,986 $ \\
			MLRSNet~\cite{qi2020mlrsnet} & Aerial/Sat RGB & Multi-Label & $ 109\,161 $ \\
			BigEarthNet~\cite{Bigearth_1} & Sat. Multispectral & Multi-Label & $ 590\,326 $ \\
			SEN12MS~\cite{sen12ms_dataset_2019} & Sat. Multispectral & Multi-Label + Region Label & $ 541\,986 $ \\
			This Work      & Sat. Multispectral & Multi-Label + Pixel-Label & $ 1\,759\,830 $ \\
			\bottomrule 
	\end{tabular}}
	\vspace{0.4em}
	\caption{Examples of current remote sensing scene benchmarks.}
	\centering
	\label{tab:dataset_comparison}
	\vspace{-1.5em}
\end{table}
\vspace{-0.8em}
\section{Experimental Results}
\label{sec:experiments}
\vspace{-0.8em}
For supervised learning tasks we provide two sets of roughly $ 880\,000 $ images, one from each grid of SeasoNet, while prohibiting sample locations from the training set to appear in the test set. 
To ensure a machine learning benchmark suitable evaluation protocol, we provide an official split for the various possible remote sensing learning tasks. Since land cover data is inherently highly imbalanced, we focused on creating a set of training / validation / test splits with low variation between the class distribution over the total dataset versus that of the individual splits, see Fig.~\ref{fig:split_classes}. In analogy to average precision thresholds, e.g.\ AP50, from instance segmentation tasks, we define three region size thresholds for our dataset, namely \textit{easy}, \textit{medium} and \textit{hard}, with $ 300\ \mathrm{px} $, $ 100\ \mathrm{px} $ and $ 0\ \mathrm{px} $ region size thresholds, respectively (Table~\ref{tab:region_th}). To control label noise introduced by small cutoff regions from neighboring patches, every region with a number of pixels below the chosen threshold is not considered during the training and evaluation process. This also controls the given multi-labels on an image level. We evaluate our dataset in land cover classification with DenseNet121~\cite{huang2017densely} and in segmentation with DeepLabV3~\cite{chen2017rethinking} using the proposed split over grid 2. All spectral bands are upsampled to the maximum patch resolution and used during training. For classification, we use a binary cross entropy loss and for segmentation the standard cross entropy loss. 
The results shown in Table~\ref{tab:experimental results} serve as a baseline. The resulting performance from current deep learning models is already producing decent accuracies for the usage in remote sensing applications. However, there is still room left for improvement on minority class and small region detection, deteriorating the average performance. Especially, the consideration of context from different regions and minority class agnostic objective functions or sampling, as in~\cite{Kossmann2020-TTM}, are promising areas of future research.

\begin{table}[b!]
	\resizebox{1\linewidth}{!}{%
		\renewcommand{\arraystretch}{0.6}   
		\begin{tabular}{m{1.6cm}m{1.6cm}m{1.6cm}m{1.6cm}m{1.6cm}}
			\toprule
			\multicolumn{1}{c}{\textbf{Sentinel-2}} & \multicolumn{1}{c}{\textit{Hard}} & \multicolumn{1}{c}{\textit{Medium}} & \multicolumn{1}{c}{\textit{Easy}}  & \multicolumn{1}{c}{\textit{Prediction}}\\
			\midrule
			
			\includegraphics[width=\samplesize\textwidth]{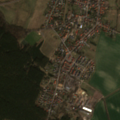} \centering
			&
			\includegraphics[width=\samplesize\textwidth]{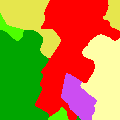} \centering
			&
			\includegraphics[width=\samplesize\textwidth]{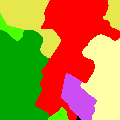} \centering
			&
			\includegraphics[width=\samplesize\textwidth]{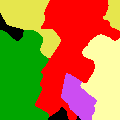} \centering
			&
			\includegraphics[width=\samplesize\textwidth]{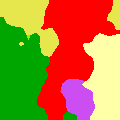} \centering
			\tabularnewline
			
			\bottomrule 
	\end{tabular}}
	\caption{Example of the region size threshold levels controlling the number of small regions flowing into the dataset's pixel and image level ground-truth annotations. Black regions do not contribute to the patch annotations and evaluation anymore. Rightmost picture: Prediction from DeepLabV3 model, trained with the \textit{hard} annotations.}
	\label{tab:region_th}
	\centering
	\vspace{-1.5em}
\end{table}

\begin{table}[h!]
	\vspace{-0.8em}
	\resizebox{1\linewidth}{!}{%
		\begin{tabular}{lcccccc}
			\toprule
			                        & \multicolumn{2}{c}{DenseNet121}               & \multicolumn{2}{c}{DenseNet121 PT}			& DeepLabV3     & DeepLabV3 PT  \\
									\cmidrule(lr){2-3}								\cmidrule(lr){4-5}
									& $\mathbf{F1}_{macro}$ & $\mathbf{F1}_{micro}$ & $\mathbf{F1}_{macro}$ & $\mathbf{F1}_{micro}$ & \textbf{mIoU} & \textbf{mIoU} \\
									
			\midrule
			\textit{hard}           & $ 60.02 $             & $ 84.14 $             & $ 59.39 $             & $ 83.07 $             & $ 47.53 $     & $ 48.69 $     \\
			\textit{medium}         & $ 61.48 $             & $ 85.06 $             & $ 60.42 $             & $ 83.46 $             & $ 47.78 $     & $ 48.95 $     \\
			\textit{easy}           & $ 61.47 $             & $ 84.10 $             & $ 59.75 $             & $ 82.11 $             & $ 48.49 $     & $ 49.66 $     \\
			$ < 300 \ \mathrm{px} $ & -                     & -                     & -                     & -                     & $ 15.98 $     & $ 16.11 $     \\
			$ < 100 \ \mathrm{px} $ & -                     & -                     & -                     & -                     & $ 10.07 $     & $ 10.13 $     \\
			\bottomrule 
		\end{tabular}}
	\vspace{0.4em}
	\caption{Results on standard classification and segmentation tasks for the SeasoNet grid 2 test set. PT: Models pre-trained on ImageNet.$\ \mathbf{F1}_{macro}$ averages over all classes equally, while $\mathbf{F1}_{micro}$ is sample weighted and thus favors correct majority class predictions. Four different models were trained on the \textit{hard} threshold image regions, thus including all small regions. Five different evaluations were carried out per model, three with the different levels of region thresholding and two specifically only on small regions below 300/100$\ \mathrm{px}$. The last two show the challenge of detecting cutoff regions.}
	\centering
	\label{tab:experimental results}
\end{table}
\vspace{-0.8em}
\section{Conclusion}
\label{sec:conclusion}
\vspace{-0.8em}
We presented a new large-scale benchmark dataset for remote sensing scene understanding comprised of $1\,759\,830$ Sentinel-2 image patches. 
It is annotated with large scale pixel level labels ($ 5 \ \mathrm{ha} $ MMU) and provides pixel synchronous examples from all four seasons, plus an additional snowy set.
SeasoNet will provide the basis for future research of deep learning models in RS, specifically in the areas of scene classification, mapping and retrieval, and domain adaption in the context of seasons.
Self-supervised learning approaches on a pixel level might now be possible and could be trained on this dataset. Experimental results show decent accuracies and the potential of SeasoNet. By providing an official evaluation protocol for various scene tasks we aim to foster the comparability of future research in this domain.
\vspace{-0.8em}
\section{Acknowledgement}
\vspace{-0.8em}
This work has been funded by the Deutsche Forschungsge-
meinschaft (DFG, German Research Foundation) - Project
number 269661170.
\vspace{-0.8em}
\bibliographystyle{IEEEbib.bst}
\bibliography{paper}

\end{document}

%% file: split_classes_grid2.tex
\begin{tikzpicture}[scale=0.7]

\definecolor{color0}{rgb}{0.12156862745098,0.466666666666667,0.705882352941177}
\definecolor{color1}{rgb}{1,0.498039215686275,0.0549019607843137}
\definecolor{color2}{rgb}{0.172549019607843,0.627450980392157,0.172549019607843}
\definecolor{color3}{rgb}{0.83921568627451,0.152941176470588,0.156862745098039}

\begin{axis}[
legend cell align={left},
legend style={fill opacity=0.8, draw opacity=1, text opacity=1, draw=white!80!black},
tick align=outside,
tick pos=left,
x grid style={white!69.0196078431373!black},
xlabel={Class index},
xmin=0.25, xmax=33.75,
xtick style={color=black},
y grid style={white!69.0196078431373!black},
ylabel={No. of images per class [\%]},
ymin=0, ymax=80.5271385444927,
ytick style={color=black},
xtick distance=1,
xticklabel style={rotate=70},
tick label style={font=\tiny},
x=0.32cm,
y=0.5cm/10
]
\draw[draw=none,fill=color0] (axis cs:0.6,0) rectangle (axis cs:0.8,3.57094933441289);
\addlegendimage{ybar,ybar legend,draw=none,fill=color0}
\addlegendentry{Total}

\draw[draw=none,fill=color0] (axis cs:1.6,0) rectangle (axis cs:1.8,47.8853138710037);
\draw[draw=none,fill=color0] (axis cs:2.6,0) rectangle (axis cs:2.8,18.7590263073505);
\draw[draw=none,fill=color0] (axis cs:3.6,0) rectangle (axis cs:3.8,6.34883451182318);
\draw[draw=none,fill=color0] (axis cs:4.6,0) rectangle (axis cs:4.8,0.125202701012311);
\draw[draw=none,fill=color0] (axis cs:5.6,0) rectangle (axis cs:5.8,0.514684309520181);
\draw[draw=none,fill=color0] (axis cs:6.6,0) rectangle (axis cs:6.8,2.89217102165859);
\draw[draw=none,fill=color0] (axis cs:7.6,0) rectangle (axis cs:7.8,0.912467277859022);
\draw[draw=none,fill=color0] (axis cs:8.6,0) rectangle (axis cs:8.8,0.487733319384016);
\draw[draw=none,fill=color0] (axis cs:9.6,0) rectangle (axis cs:9.8,2.60332918644399);
\draw[draw=none,fill=color0] (axis cs:10.6,0) rectangle (axis cs:10.8,11.0371696229363);
\draw[draw=none,fill=color0] (axis cs:11.6,0) rectangle (axis cs:11.8,75.6763333917082);
\draw[draw=none,fill=color0] (axis cs:12.6,0) rectangle (axis cs:12.8,1.47525398749565);
\draw[draw=none,fill=color0] (axis cs:13.6,0) rectangle (axis cs:13.8,4.88222303593238);
\draw[draw=none,fill=color0] (axis cs:14.6,0) rectangle (axis cs:14.8,76.5215937701137);
\draw[draw=none,fill=color0] (axis cs:15.6,0) rectangle (axis cs:15.8,51.2536190517345);
\draw[draw=none,fill=color0] (axis cs:16.6,0) rectangle (axis cs:16.8,52.2111183637451);
\draw[draw=none,fill=color0] (axis cs:17.6,0) rectangle (axis cs:17.8,37.1996442924171);
\draw[draw=none,fill=color0] (axis cs:18.6,0) rectangle (axis cs:18.8,3.84739598851001);
\draw[draw=none,fill=color0] (axis cs:19.6,0) rectangle (axis cs:19.8,1.23042073111099);
\draw[draw=none,fill=color0] (axis cs:20.6,0) rectangle (axis cs:20.8,16.7117745123235);
\draw[draw=none,fill=color0] (axis cs:21.6,0) rectangle (axis cs:21.8,0.357413341763573);
\draw[draw=none,fill=color0] (axis cs:22.6,0) rectangle (axis cs:22.8,0.0774414526697401);
\draw[draw=none,fill=color0] (axis cs:23.6,0) rectangle (axis cs:23.8,0.345018160646096);
\draw[draw=none,fill=color0] (axis cs:24.6,0) rectangle (axis cs:24.8,2.14095481558472);
\draw[draw=none,fill=color0] (axis cs:25.6,0) rectangle (axis cs:25.8,1.45330655670966);
\draw[draw=none,fill=color0] (axis cs:26.6,0) rectangle (axis cs:26.8,0.292025918437434);
\draw[draw=none,fill=color0] (axis cs:27.6,0) rectangle (axis cs:27.8,1.263285018661);
\draw[draw=none,fill=color0] (axis cs:28.6,0) rectangle (axis cs:28.8,3.77268375003127);
\draw[draw=none,fill=color0] (axis cs:29.6,0) rectangle (axis cs:29.8,6.68964513392481);
\draw[draw=none,fill=color0] (axis cs:30.6,0) rectangle (axis cs:30.8,0.509339598396132);
\draw[draw=none,fill=color0] (axis cs:31.6,0) rectangle (axis cs:31.8,0.265643514591061);
\draw[draw=none,fill=color0] (axis cs:32.6,0) rectangle (axis cs:32.8,5.82266475924919);
\draw[draw=none,fill=color1] (axis cs:0.8,0) rectangle (axis cs:1,3.54431202424183);
\addlegendimage{ybar,ybar legend,draw=none,fill=color1}
\addlegendentry{Train}

\draw[draw=none,fill=color1] (axis cs:1.8,0) rectangle (axis cs:2,47.8631603746761);
\draw[draw=none,fill=color1] (axis cs:2.8,0) rectangle (axis cs:3,18.8397390590854);
\draw[draw=none,fill=color1] (axis cs:3.8,0) rectangle (axis cs:4,6.36039417351108);
\draw[draw=none,fill=color1] (axis cs:4.8,0) rectangle (axis cs:5,0.126895924219899);
\draw[draw=none,fill=color1] (axis cs:5.8,0) rectangle (axis cs:6,0.518307296109446);
\draw[draw=none,fill=color1] (axis cs:6.8,0) rectangle (axis cs:7,2.85848910986002);
\draw[draw=none,fill=color1] (axis cs:7.8,0) rectangle (axis cs:8,0.921417139886102);
\draw[draw=none,fill=color1] (axis cs:8.8,0) rectangle (axis cs:9,0.499134800516682);
\draw[draw=none,fill=color1] (axis cs:9.8,0) rectangle (axis cs:10,2.63118130194243);
\draw[draw=none,fill=color1] (axis cs:10.8,0) rectangle (axis cs:11,11.0765031317784);
\draw[draw=none,fill=color1] (axis cs:11.8,0) rectangle (axis cs:12,75.6749774560698);
\draw[draw=none,fill=color1] (axis cs:12.8,0) rectangle (axis cs:13,1.51348980039482);
\draw[draw=none,fill=color1] (axis cs:13.8,0) rectangle (axis cs:14,4.8540534392695);
\draw[draw=none,fill=color1] (axis cs:14.8,0) rectangle (axis cs:15,76.5237665829901);
\draw[draw=none,fill=color1] (axis cs:15.8,0) rectangle (axis cs:16,51.1463690055486);
\draw[draw=none,fill=color1] (axis cs:16.8,0) rectangle (axis cs:17,52.1867206096204);
\draw[draw=none,fill=color1] (axis cs:17.8,0) rectangle (axis cs:18,37.1608458645089);
\draw[draw=none,fill=color1] (axis cs:18.8,0) rectangle (axis cs:19,3.85188434760709);
\draw[draw=none,fill=color1] (axis cs:19.8,0) rectangle (axis cs:20,1.2345137416425);
\draw[draw=none,fill=color1] (axis cs:20.8,0) rectangle (axis cs:21,16.6724346632221);
\draw[draw=none,fill=color1] (axis cs:21.8,0) rectangle (axis cs:22,0.35339133825644);
\draw[draw=none,fill=color1] (axis cs:22.8,0) rectangle (axis cs:23,0.0760400672662133);
\draw[draw=none,fill=color1] (axis cs:23.8,0) rectangle (axis cs:24,0.345104920669738);
\draw[draw=none,fill=color1] (axis cs:24.8,0) rectangle (axis cs:25,2.15463105131892);
\draw[draw=none,fill=color1] (axis cs:25.8,0) rectangle (axis cs:26,1.40446654155801);
\draw[draw=none,fill=color1] (axis cs:26.8,0) rectangle (axis cs:27,0.291811882072904);
\draw[draw=none,fill=color1] (axis cs:27.8,0) rectangle (axis cs:28,1.27773309611432);
\draw[draw=none,fill=color1] (axis cs:28.8,0) rectangle (axis cs:29,3.78949249754251);
\draw[draw=none,fill=color1] (axis cs:29.8,0) rectangle (axis cs:30,6.70419926397114);
\draw[draw=none,fill=color1] (axis cs:30.8,0) rectangle (axis cs:31,0.51359541159936);
\draw[draw=none,fill=color1] (axis cs:31.8,0) rectangle (axis cs:32,0.267115108089006);
\draw[draw=none,fill=color1] (axis cs:32.8,0) rectangle (axis cs:33,5.80406684376853);
\draw[draw=none,fill=color2] (axis cs:1,0) rectangle (axis cs:1.2,3.73476995817421);
\addlegendimage{ybar,ybar legend,draw=none,fill=color2}
\addlegendentry{Val}

\draw[draw=none,fill=color2] (axis cs:2,0) rectangle (axis cs:2.2,47.7950536461175);
\draw[draw=none,fill=color2] (axis cs:3,0) rectangle (axis cs:3.2,18.7863702491362);
\draw[draw=none,fill=color2] (axis cs:4,0) rectangle (axis cs:4.2,6.42162211311148);
\draw[draw=none,fill=color2] (axis cs:5,0) rectangle (axis cs:5.2,0.131842153118749);
\draw[draw=none,fill=color2] (axis cs:6,0) rectangle (axis cs:6.2,0.445535551918531);
\draw[draw=none,fill=color2] (axis cs:7,0) rectangle (axis cs:7.2,2.94371703946172);
\draw[draw=none,fill=color2] (axis cs:8,0) rectangle (axis cs:8.2,0.8978905255501);
\draw[draw=none,fill=color2] (axis cs:9,0) rectangle (axis cs:9.2,0.440989270776505);
\draw[draw=none,fill=color2] (axis cs:10,0) rectangle (axis cs:10.2,2.52773231496636);
\draw[draw=none,fill=color2] (axis cs:11,0) rectangle (axis cs:11.2,11.1793053282415);
\draw[draw=none,fill=color2] (axis cs:12,0) rectangle (axis cs:12.2,75.350063647936);
\draw[draw=none,fill=color2] (axis cs:13,0) rectangle (axis cs:13.2,1.41503000545554);
\draw[draw=none,fill=color2] (axis cs:14,0) rectangle (axis cs:14.2,4.86679396253864);
\draw[draw=none,fill=color2] (axis cs:15,0) rectangle (axis cs:15.2,76.3922985997454);
\draw[draw=none,fill=color2] (axis cs:16,0) rectangle (axis cs:16.2,51.4229859974541);
\draw[draw=none,fill=color2] (axis cs:17,0) rectangle (axis cs:17.2,51.9685397344972);
\draw[draw=none,fill=color2] (axis cs:18,0) rectangle (axis cs:18.2,36.7305419167121);
\draw[draw=none,fill=color2] (axis cs:19,0) rectangle (axis cs:19.2,3.8916166575741);
\draw[draw=none,fill=color2] (axis cs:20,0) rectangle (axis cs:20.2,1.27182214948172);
\draw[draw=none,fill=color2] (axis cs:21,0) rectangle (axis cs:21.2,16.5052736861247);
\draw[draw=none,fill=color2] (axis cs:22,0) rectangle (axis cs:22.2,0.371658483360611);
\draw[draw=none,fill=color2] (axis cs:23,0) rectangle (axis cs:23.2,0.0761502091289325);
\draw[draw=none,fill=color2] (axis cs:24,0) rectangle (axis cs:24.2,0.426213857064921);
\draw[draw=none,fill=color2] (axis cs:25,0) rectangle (axis cs:25.2,2.18562465902891);
\draw[draw=none,fill=color2] (axis cs:26,0) rectangle (axis cs:26.2,1.49799963629751);
\draw[draw=none,fill=color2] (axis cs:27,0) rectangle (axis cs:27.2,0.264820876523004);
\draw[draw=none,fill=color2] (axis cs:28,0) rectangle (axis cs:28.2,1.25818330605565);
\draw[draw=none,fill=color2] (axis cs:29,0) rectangle (axis cs:29.2,3.70635570103655);
\draw[draw=none,fill=color2] (axis cs:30,0) rectangle (axis cs:30.2,6.6012002182215);
\draw[draw=none,fill=color2] (axis cs:31,0) rectangle (axis cs:31.2,0.479632660483724);
\draw[draw=none,fill=color2] (axis cs:32,0) rectangle (axis cs:32.2,0.223904346244772);
\draw[draw=none,fill=color2] (axis cs:33,0) rectangle (axis cs:33.2,6.13861611202037);
\draw[draw=none,fill=color3] (axis cs:1.2,0) rectangle (axis cs:1.4,3.58220832741225);
\addlegendimage{ybar,ybar legend,draw=none,fill=color3}
\addlegendentry{Test}

\draw[draw=none,fill=color3] (axis cs:2.2,0) rectangle (axis cs:2.4,48.0079579366207);
\draw[draw=none,fill=color3] (axis cs:3.2,0) rectangle (axis cs:3.4,18.4629813841125);
\draw[draw=none,fill=color3] (axis cs:4.2,0) rectangle (axis cs:4.4,6.27199090521529);
\draw[draw=none,fill=color3] (axis cs:5.2,0) rectangle (axis cs:5.4,0.115958505044763);
\draw[draw=none,fill=color3] (axis cs:6.2,0) rectangle (axis cs:6.4,0.536592297854199);
\draw[draw=none,fill=color3] (axis cs:7.2,0) rectangle (axis cs:7.4,2.98422623276965);
\draw[draw=none,fill=color3] (axis cs:8.2,0) rectangle (axis cs:8.4,0.888446781298849);
\draw[draw=none,fill=color3] (axis cs:9.2,0) rectangle (axis cs:9.4,0.471223532755436);
\draw[draw=none,fill=color3] (axis cs:10.2,0) rectangle (axis cs:10.4,2.54369759840841);
\draw[draw=none,fill=color3] (axis cs:11.2,0) rectangle (axis cs:11.4,10.8284780446213);
\draw[draw=none,fill=color3] (axis cs:12.2,0) rectangle (axis cs:12.4,75.8442518118516);
\draw[draw=none,fill=color3] (axis cs:13.2,0) rectangle (axis cs:13.4,1.37160721898536);
\draw[draw=none,fill=color3] (axis cs:14.2,0) rectangle (axis cs:14.4,4.98848941310217);
\draw[draw=none,fill=color3] (axis cs:15.2,0) rectangle (axis cs:15.4,76.5786556771351);
\draw[draw=none,fill=color3] (axis cs:16.2,0) rectangle (axis cs:16.4,51.5441239164417);
\draw[draw=none,fill=color3] (axis cs:17.2,0) rectangle (axis cs:17.4,52.4177916725878);
\draw[draw=none,fill=color3] (axis cs:18.2,0) rectangle (axis cs:18.4,37.569987210459);
\draw[draw=none,fill=color3] (axis cs:19.2,0) rectangle (axis cs:19.4,3.80957794514708);
\draw[draw=none,fill=color3] (axis cs:20.2,0) rectangle (axis cs:20.4,1.19539576524087);
\draw[draw=none,fill=color3] (axis cs:21.2,0) rectangle (axis cs:21.4,16.9526786983089);
\draw[draw=none,fill=color3] (axis cs:22.2,0) rectangle (axis cs:22.4,0.364359812420065);
\draw[draw=none,fill=color3] (axis cs:23.2,0) rectangle (axis cs:23.4,0.082989910473213);
\draw[draw=none,fill=color3] (axis cs:24.2,0) rectangle (axis cs:24.4,0.304106863720335);
\draw[draw=none,fill=color3] (axis cs:25.2,0) rectangle (axis cs:25.4,2.07076879351997);
\draw[draw=none,fill=color3] (axis cs:26.2,0) rectangle (axis cs:26.4,1.60181895694188);
\draw[draw=none,fill=color3] (axis cs:27.2,0) rectangle (axis cs:27.4,0.306380559897684);
\draw[draw=none,fill=color3] (axis cs:28.2,0) rectangle (axis cs:28.4,1.21529060679267);
\draw[draw=none,fill=color3] (axis cs:29.2,0) rectangle (axis cs:29.4,3.74705130027);
\draw[draw=none,fill=color3] (axis cs:30.2,0) rectangle (axis cs:30.4,6.682961489271);
\draw[draw=none,fill=color3] (axis cs:31.2,0) rectangle (axis cs:31.4,0.50930794372602);
\draw[draw=none,fill=color3] (axis cs:32.2,0) rectangle (axis cs:32.4,0.281369901946852);
\draw[draw=none,fill=color3] (axis cs:33.2,0) rectangle (axis cs:33.4,5.72971436691772);
\end{axis}

\end{tikzpicture}